\documentclass[letterpaper, 10 pt, conference]{ieeeconf}  % Comment this line out
\IEEEoverridecommandlockouts    
\usepackage{cite}
\newcommand{\eat}[1]{}
\usepackage{booktabs}
\usepackage{color}
\usepackage [autostyle, english = american]{csquotes}
\MakeOuterQuote{"}
\usepackage{xcolor}

\usepackage{multicol}

\usepackage{mathtools}
    
\usepackage{amsmath}
\usepackage{amstext}% assumes amsmath package installed
\usepackage{amssymb}
\usepackage{amsfonts}
\usepackage{float}

\usepackage{amsthm}  %for making assumptions bold

\usepackage{todonotes}

\makeatletter

\newcommand{\Rmnum}[1]{\expandafter\@slowromancap\romannumeral #1@}
\usepackage{savesym}

\usepackage{algorithm}
\usepackage{algorithmicx} % <===========================================
\usepackage{algpseudocode} %
\savesymbol{AND}
\usepackage[group-separator={,},group-minimum-digits={3}]{siunitx}

\usepackage{graphicx} % for pdf, bitmapped graphics files
\usepackage{epsfig} % for postscript graphics files

\usepackage{times} % assumes new font selection scheme installed
\usepackage{amsmath} % assumes amsmath package installed
\usepackage{amssymb}  % assumes amsmath package installed
\usepackage{comment}
\makeatletter
\let\NAT@parse\undefined
\makeatother
\usepackage{hyperref}
\hypersetup{
   colorlinks=true,
    linkcolor= blue,
    allcolors=blue,
    citecolor = blue,
    filecolor=black,      
    urlcolor=blue,
    }
\usepackage{mathrsfs}
\title{\LARGE \bf
A Digital Smart City for Emerging Mobility Systems

}

\author{Raymond M. Zayas, Logan E. Beaver, \emph{IEEE Member}, Behdad Chalaki, \emph{IEEE Member},\\Heeseung Bang, \emph{IEEE Student Member}, Andreas A. Malikopoulos, \emph{IEEE Senior Member}
\thanks{This work was supported by NSF under Grants CNS-2149520 and CMMI-2219761.}
\thanks{R.M. Zayas, H. Bang, and A.A. Malikopoulos are with the Department of Mechanical Engineering, University of Delaware, Newark, DE 19716 USA (emails: \texttt{\{rayzayas;heeseung;andreas\}@udel.edu})}
\thanks{B. Chalaki is with the Honda Research Institute, Ann Arbor, MI 48103, USA (email: \texttt{bchalaki@udel.edu})}
\thanks{L.E. Beaver is with the Division of Systems Engineering, Boston University, Brookline, MA 02246 USA (email: \texttt{lebeaver@bu.edu})}}
\begin{document}

\maketitle
\thispagestyle{empty}
\pagestyle{empty}

%%%%%%%%%%%%%%%%%%%%%%%%%%%%%%%%%%%%%%%%%%%%%%%%%%%%%%%%%%%%%%%%%%%%%%%%%%%%%%%%
\begin{abstract}
The increasing demand for emerging mobility systems with connected and automated vehicles has imposed the necessity for quality testing environments to support their development.
In this paper, we introduce a Unity-based virtual simulation environment for emerging mobility systems, called the Information and Decision Science Lab's Scaled Smart Digital City (IDS $3$D City), intended to operate alongside its physical peer and its established control framework.
By utilizing the Robot Operation System, AirSim, and Unity, we constructed a simulation environment capable of iteratively designing experiments significantly faster than it is possible in a physical testbed.
This environment provides an intermediate step to validate the effectiveness of our control algorithms prior to their implementation in the physical testbed.
The IDS $3$D City also enables us to demonstrate that our control algorithms work independently of the underlying vehicle dynamics, as the vehicle dynamics introduced by AirSim operate at a different scale than our scaled smart city.
Finally, we demonstrate the behavior of our digital environment by performing an experiment in both the virtual and physical environments and comparing their outputs. 

\end{abstract}

\section{Introduction}
\PARstart{O}{ver} the last decade, the growing population in urban areas, without a corresponding increase in road capacity, has led to traffic congestion, increased delays, and environmental concerns \cite{Schrank2019}. 
Integrating communication technologies along with computational capabilities into connected and automated vehicles (CAVs) has the potential to revolutionize our overwhelmed transportation systems. Through these advancements, our transportation system will transition into an emerging mobility system, in which CAVs can make better operational decisions\textemdash leading to improvements in passengers safety as well as a significant reduction of energy consumption, greenhouse gas emissions, and travel delays \cite{zhao2019enhanced,Melo2017a,ersal2020connected,Mahbub2019ACC,Wadud2016,chalaki2020TCST}.

Rigorous evaluation of the performance of CAVs requires a broad spectrum of testing, ranging from numerical simulation to real-world public roads. Recently, the emergence of scaled cities has received significant global attention as a more sustainable CAV testing solution \cite{paull2017duckietown,hyldmar2019fleet,fok2012platform,Beaver2020DemonstrationCity,chalaki2021CSM}. These closed-test facilities use robotic cars to ensure safety, complete control of the test-environment variables, and quick, repeatable experiments. A key intermediate step before testing these new technologies in a scaled environment is to use high-fidelity simulations to gather preliminary information about the system's performance in an idealized environment \cite{Wang2010_its}.

%\todo[inline]{Maybe here we can emphasize that this is system-level and not vehicle-level, two reviewers wanted more specifics on how the CAVs do perception, which we don't care about.}

Several research efforts have been reported in the literature on creating a digital version of the real environment using physics-based simulation software.
Zhang and Masoud \cite{zhang2020v2xsim} used Gazebo to create a virtual environment to test CAVs due to its ability to capture microscopic vehicle movement. 
The authors selected Gazebo, rather than a game engine, due to concerns about rigorously replicating the full dynamics of an individual vehicle. 
In other efforts, a simulation framework for CAVs has been linked to the robot operating system (ROS) and  game-engine platform Unity \cite{hussein2018ros,tsai2017virtualreality,mizuchi2017interaction}.
Tsai et al. \cite{tsai2017virtualreality} demonstrated the validity of hardware-in-the-loop simulation utilizing the ROS Unity link. 
Mizuchi et al. \cite{mizuchi2017interaction} introduced virtual reality for multiple users into the environment using Unity, and Yang et al. \cite{yang2016unityproduction} modeled an existing environment within Unity to validate simulated sensors in a variety of weather and lighting conditions.

In this paper, we introduce the Information and Decision Science Laboratory's Scaled Smart Digital City (IDS 3D City) in Unity, a full-scale digital recreation of the Information and Decision Science Lab's Scaled Smart City (IDS$^3$C) physical testbed \cite{Beaver2020DemonstrationCity,chalaki2021CSM}. We are particularly interested in how the collective behavior of the CAVs influences the transportation network, and our focus is not on the perception and low-level control of individual vehicles.
IDS$^3$C is a $1$:$25$ scaled testbed spanning over $400$ square feet, and it is capable of replicating real-world traffic scenarios using up to $50$ ground and $10$ aerial vehicles.
Our digital replica can communicate with the central mainframe computer using the user datagram protocol (UDP), allowing users and potential collaborators to evaluate the behavior of their algorithms before running a physical experiment in the IDS$^3$C.
Using IDS 3D City, we are also able to rapidly iterate the design of our experiments before deploying them on the physical city.

While other virtual environments have been created to test the performance of individual CAV, to the best of our knowledge, the environment we report in this paper is the first one that is capable of analyzing a transportation network at a system level.
IDS 3D City also facilitates the investigation of different traffic scenarios, such as coordination of CAVs in the presence of human-driven vehicles. Another benefit of IDS 3D City is that it allows users to test their control algorithms in the physical environment after validating them in the virtual environment without further changes. 
This is particularly important; unlike existing simulators, e.g., CARLA, the IDS 3D City directly interfaces with the control software that operates the IDS$^3$C.

The remainder of the paper proceeds as follows.
In Section \ref{sec:simulation}, we introduce IDS 3D City and elaborate on the  different features and their interactions. In Section \ref{sec:experiment}, we present a coordination problem of CAVs at a roundabout, and compare the results from IDS 3D City and IDS$^3$C. Finally, we draw concluding remarks and propose some directions for future research in Section \ref{sec:conclusion}.

\section{Digital Simulation Environment} \label{sec:simulation}
\begin{figure*}[ht]
    \centering
    \includegraphics[width=\linewidth]{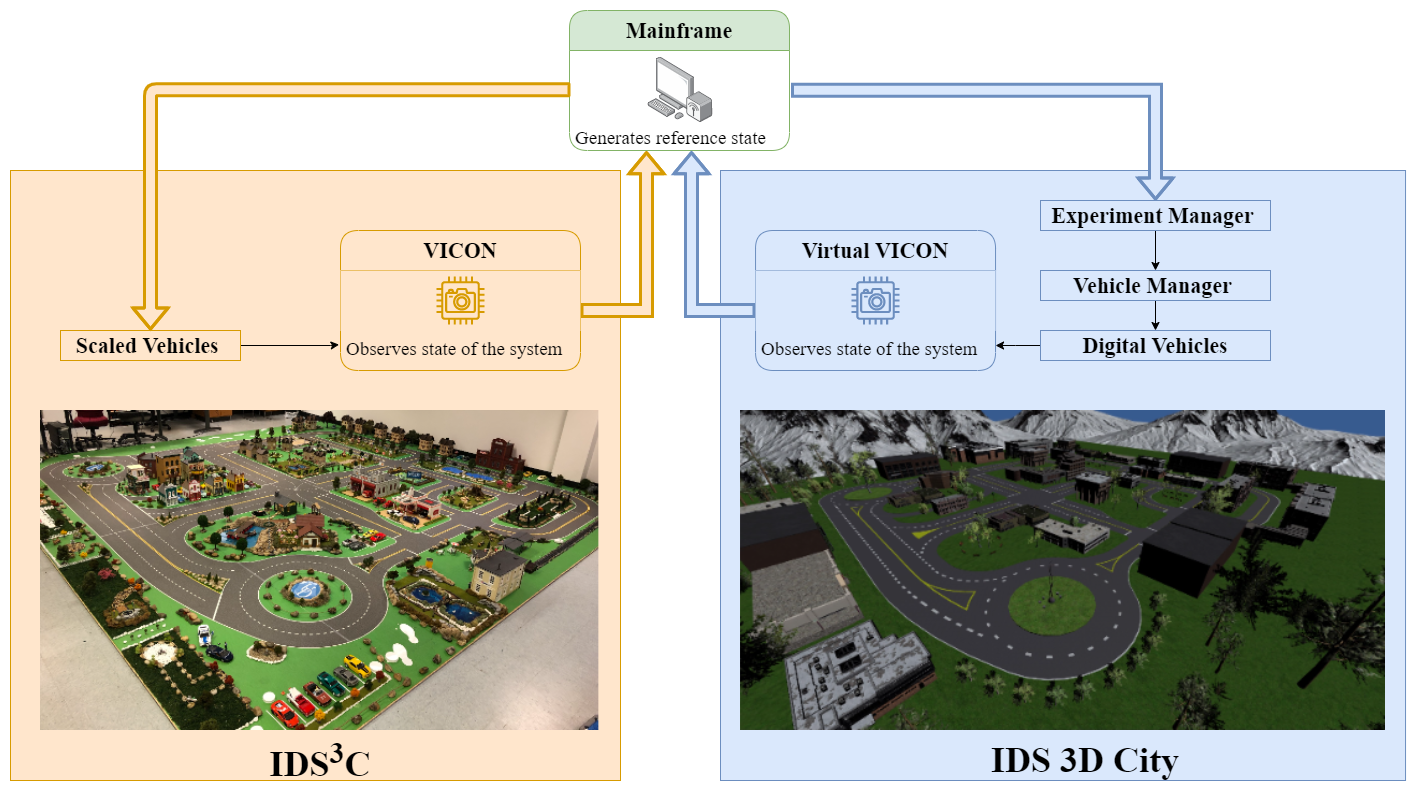}
    \caption{Comparison of the physical and virtual city environments. The mainframe computer can switch between physical and virtual experiment seamlessly.}
    \label{fig:commGraph}
\end{figure*}

The IDS $3$D City integrates seamlessly the current control framework used in its physical counterpart, IDS$^3$C. A schematic of the communication structure between the IDS $3$D City and IDS$^3$C is shown in Fig. \ref{fig:commGraph}.
During a physical experiment, a central mainframe computer runs a custom C++ application that generates a separate thread for each CAV in the experiment.
Each physical CAV in IDS$^3$C receives a desired trajectory from the mainframe computer over WiFi, and the position and orientation of each CAV are fed back to the mainframe computer through a VICON motion capture system.

To imitate the behavior of IDS$^3$C, we send trajectory data over a local UDP socket from mainframe to the IDS $3$D City application.
This trajectory data consists of the desired state of each CAV in the simulation.
After each physics update, the position and orientation of each CAV in the IDS $3$D City are broadcast through ROS to a node that mimics the format of VICON measurements.
This information is accessed by the mainframe computer, which updates the CAVs' states, executes the control algorithm, and sends new commands over UDP.
A major consequence of this design is that we can seamlessly switch between running any individual car in the physical or virtual environment with minimal changes to our input files.
The IDS $3$D City is also capable of replaying experimental data, allowing users to directly control a vehicle, and streaming a live feed of the virtual cameras attached to each vehicle.
In the following subsections, we review the three major aspects of our simulation environment: the Unity game engine, Microsoft AirSim, and ROS\#.

%%%%%%%
%UNITY%
%%%%%%%

\subsection{Unity Game Engine}
\label{sec:unity}
We built a majority of the IDS $3$D City using Unity, a free and highly-customizable game engine with built-in physics and a C\# scripting framework; for a brief history of the Unity game engine, see \cite{hussein2018ros}.
We selected Unity over existing simulation packages, such as Gazebo, as it is easy to deploy and performs well on a variety of platforms.
Unlike Zhang and Masoud \cite{zhang2020v2xsim}, our interest lies in the system-level coordination of CAVs, not the particular dynamics of any individual CAV. 
Unity also relies heavily on the entity-component paradigm of software design, which grants us incredible flexibility in the design and control of vehicles in the virtual environment.
The built-in Nvidia PhysX engine is open-source, which provides us the ability to modify the physics of the experiment when necessary. Unity is capable of building an executable for Windows, Mac, Linux, and mobile devices, which ensures that the simulation will run natively on all available hardware.
Unity's graphical settings are also configurable per device, allowing weaker hardware to access the IDS $3$D City, while more powerful hardware can produce high-fidelity videos and screenshots.
Furthermore, Unity allows us to explore more accurate mixed-traffic scenarios with built-in virtual reality support.

As a first step to creating the IDS $3$D City, we reconstructed the IDS$^3$C's road network at full scale and placed environmental decorations within Unity.
The road network was defined in CAD files, which defines each road segment as either a straight line or arc segment. 
To handle the simulation logic, we created two manager scripts.
The \textit{Experiment Manager} is the primary manager, which controls the experiment clock used for data collection.
It also stores the initial conditions of all vehicles, this ensures that an experiment can be repeated without restarting the simulation software.  
The secondary manager script is the \textit{Vehicle Manager}, which handles all of the vehicles. 
The vehicle manager spawns each vehicle at its initial position, and if two vehicles overlap, the vehicle manager places the second one behind the first to avoid infeasible initial conditions.
The vehicle manager also passes information about the vehicles to the user interface (UI) and data logging tools.

\begin{figure}[ht]
    \centering
    \includegraphics[width=0.8\linewidth]{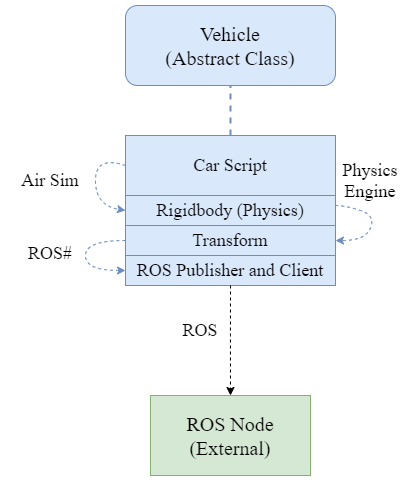}
    \caption{A diagram showing the different components that make up a single vehicle in the Unity simulation.
    The dashed arrows denote communication between different subsystems of the car script, which is a child of the abstract vehicle class.
    }
    \label{fig:prefab}
\end{figure}

To initialize vehicles into the environment, we use Unity's prefab system which allows us to configure each vehicle based on the initialization data sent from the mainframe computer. 
For each vehicle, the initialization data includes the control algorithm name, controller parameters, the initial state, and vehicle appearance.
We implemented the vehicles as an abstract class, thus the vehicle manager is flexible enough to initialize and coordinate any additional vehicle types that we may add in the future.
A schematic of the key components in our vehicle prefab is presented in Fig. \ref{fig:prefab}, and the behavior of the AirSim and ROS\# components are explained in the relevant sections that follow.

We use the passenger car as the main vehicle type in our simulation.
This is controlled by a custom car script, which is a child of the abstract vehicle class.
The car script takes a timestamped waypoint as input, which consists of a desired position in $\mathbb{R}^2$, an orientation in $\mathbb{R}$, and a speed in $\mathbb{R}$.
This information is passed to a low-level tracking controller to generate a steering angle and throttle command.
The steering angle is computed using a modified Stanley controller \cite{thrun2006stanley},
\begin{align} \label{eq:stanley}
    \delta(t) = &\big(\psi(t) - k_a\cdot v(t)\cdot\dot{\psi}(t)\big) \notag\\
    & + \arctan\Big\{ \frac{k_e y_e(t)}{k_s + v(t)} \Big\} - k_y\big(\dot{\psi}(t) - \dot{\psi}_d\big),
\end{align}
where $\delta(t)$ is the steering angle, $\psi(t)$ is current yaw angle, $\psi_d(t)$ is the desired yaw angle, $v(t)$ is the current speed, $y_e(t)$ is the lateral tracking error, $k_a, k_e, k_y$ are proportional tracking constants, and $k_s$ is a small constant that ensures the controller can operate at low speeds.
The throttle command is generated through a feedforward-feedback controller, i.e., the desired position is tracked using PID control, and we compensate for the vehicle's speed at that point with a feedforward term in the control loop \cite{Spong2004RobotEdition}.
The throttle command is sent through a second layer of the controller 
where it is translated into gas, brake, and handbrake inputs (formally defined in the next section).
Finally, the steering angle and throttle commands are sent to the AirSim controller, which updates the state of the vehicle using its own dynamic model.

\begin{figure}[ht]
    \centering
    \includegraphics[width=\linewidth]{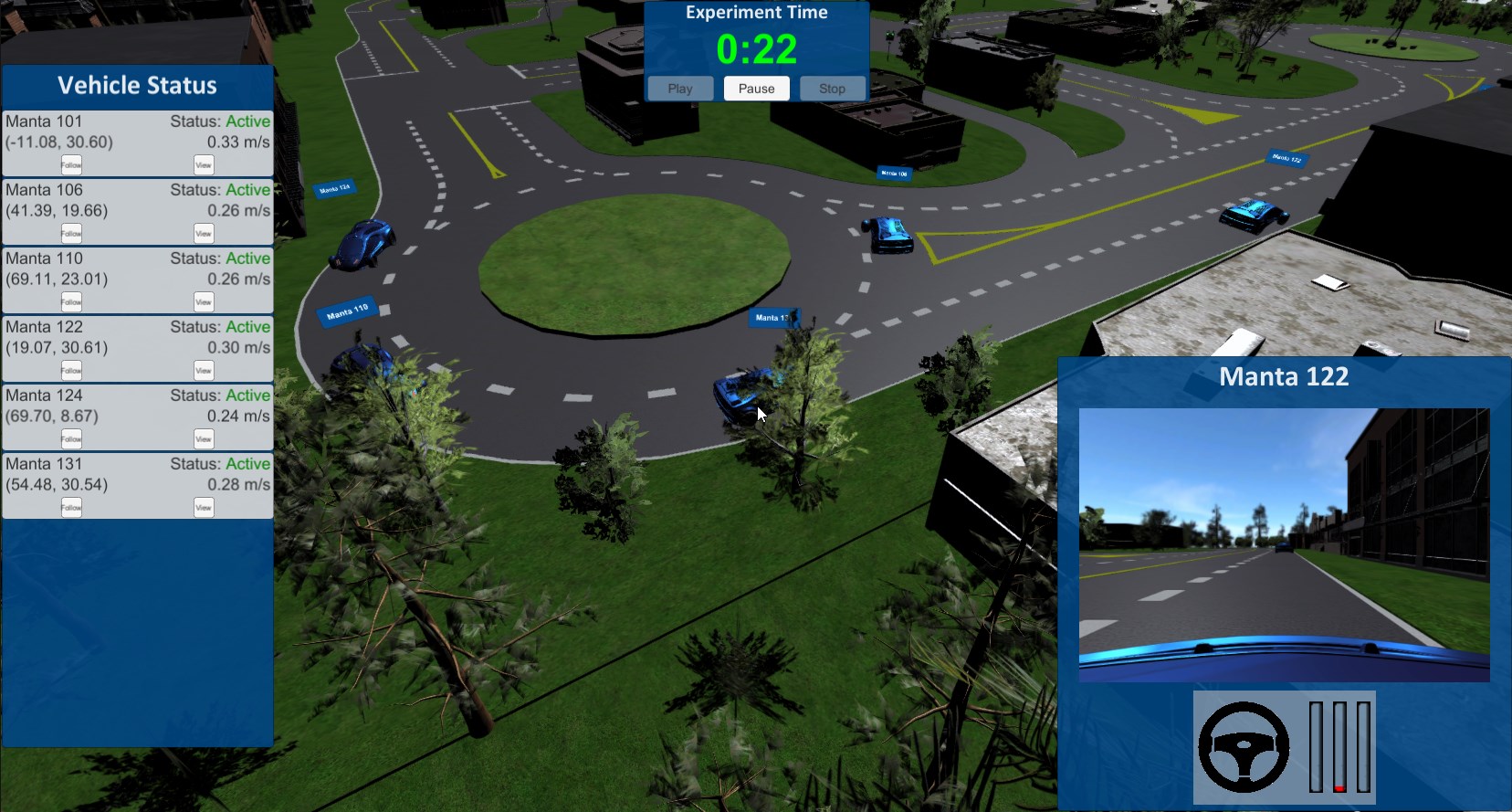}
    \caption{The digital simulation UI during one run of the experiment. The vehicle UI nodes are on the left, the experiment controls on top, and the preview panel on the right.}
    \label{fig:city}
\end{figure}

The final major component within Unity is the UI, which is visible in Fig. \ref{fig:city}. 
The UI displays information about the current experiment and CAVs in a human-readable format. The UI includes all of the relevant information about each vehicle, including the vehicle's ID, status, current position, and speed.
We also included buttons that allow the user to open a preview panel for any vehicle.
The preview panel contains a live feed of the camera attached to the CAV, as well as the current steering angle, gas, brake, and handbrake commands.

%%%%%%%%
%/UNITY%
%%%%%%%%

%%%%%%%%
%AIRSIM%
%%%%%%%%

\subsection{AirSim}
\label{airsim}
To model the dynamics and sensors of each vehicle, we included Microsoft AirSim's work-in-progress Unity module\footnote{AirSim: \url{https://microsoft.github.io/AirSim/Unity/}}.
We accomplished this by using the AirLib wrapper plugin, which allows us access to AirSim's C++ API while maintaining the Unity code base in C\#.
Our vehicle prefabs (Fig. \ref{fig:prefab}) are based on the prefabs contained in AirSim.
AirSim provides convenient code packages, for vehicles and drones, that model physically accurate behavior while being fully configurable. 
Configurable variables include motor torque, steering angle limits, weight, and aerodynamic drag.
This allows us to validate our approaches to CAV coordination on a variety of vehicles, and further helps us demonstrate that our control algorithms are independent of the underlying vehicle dynamics.
Another major feature of AirSim is its sensor suite. 
Namely, each vehicle is equipped with an RGB camera to collect qualitative data and to give visual feedback to a human operator. 

We made several modifications to the AirSim source code, both to fix undesirable behaviors and to customize the vehicles for our use case.
We fixed apparent bugs in the braking behavior, one where extreme braking would occur, and another where the brakes would lock and be unable to move.
Finally, our low-level tracking controller outputs a normalized throttle command $u_d(t) \in [-1, 1]$;
however, the AirSim controller expects three input variables, gas, brake, and handbrake.
We map the desired throttle to these variables using an intermediate layer,
\begin{align}
    h(t) &= \begin{cases}
    1 & \text{ if } u_d(t) \leq -0.5, \\
    0 & \text{ otherwise},
    \end{cases} \\
    b(t) &= \max\big\{0, -u_d(t)\big\} \cdot \big(1 - h(t)\big), \\
    g(t) &= \max\big\{0, \,\,\,\,u_d(t)\big\} \cdot \big(1 - h(t)\big),
\end{align}
where $h(t) \in \{0, 1\}$ is the handbrake, $b(t)\in[0,1]$ is the brake command, and $g(t)\in[0,1]$ is the gas command.
This results in the AirSim controller tracking the desired speed, and the vehicle only triggers the handbrake when a sufficiently large deceleration is requested.
It also guarantees that the vehicle will stop, rather than shifting into reverse, if it overshoots its current waypoint.

%%%%%%%%%
%/AIRSIM%
%%%%%%%%%

%%%%%
%ROS%
%%%%%

\subsection{ROS Framework}
ROS provides a flexible framework for robotics software, particularly through its standardized communication protocols.
These protocols give separate software components the ability to exchange information reliably, while providing access to a wide suite of debugging tools.
To introduce ROS functionality into Unity, we integrated Siemens's open-source ROS\# package\footnote{ROS\#: \url{https://github.com/siemens/ros-sharp}}.
In the IDS$^3$C, we use ROS to access VICON motion capture data and determine the state of each vehicle in real time.
In the IDS $3$D City, we use ROS\# to mimic the VICON ROS topic by attaching two ROS-specific components called publisher and client to the vehicle prefab. 

The publisher component captures the position and orientation data of the vehicle.
This information is composed into ROS messages to be published as a timestamped transform message.
The client component connects to a ROS server that runs on the mainframe computer. 
The client streams the state data of each vehicle to the ROS server, which the server broadcasts in the
same format as the VICON motion capture system.
This setup also enables us to control virtual and physical vehicles simultaneously and have access to the state information of all vehicles in real time. 

%%%%%%
%/ROS%
%%%%%%

\section{Virtual and Physical Experiment} \label{sec:experiment}
\begin{figure}[t]
    \centering
    \includegraphics[width=0.7\linewidth]{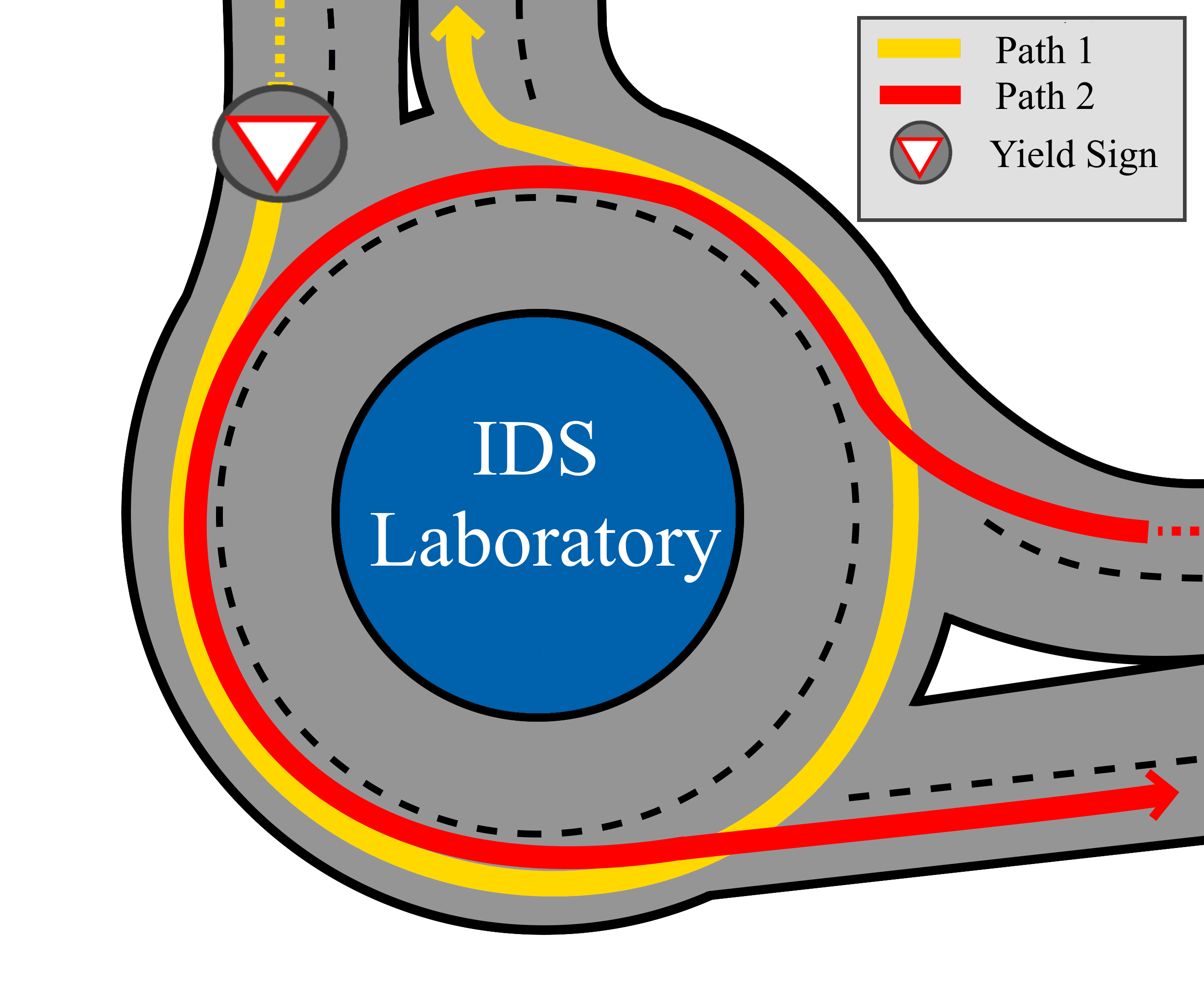}
    \caption{A schematic of the roundabout scenario showing the two paths and the yield sign location.}
    \label{fig:Roundabout}
    \vspace{-8pt}
\end{figure}

To demonstrate the capabilities of the IDS $3$D City, we consider a scenario of homogeneous human-driven vehicles operating in a single-lane roundabout, depicted in Fig. \ref{fig:Roundabout}.
We consider $N=6$ vehicles entering the roundabout in two groups of $3$, one from the northern entry and one from the eastern entry.
Our approach to planning trajectories for each vehicle $i\in\{1, 2, \dots, N\}$ considers double integrator dynamics,
\begin{align}
    \dot{p}_i(t) &= v_i(t), \\
    \dot{v}_i(t) &= u_i(t),
\end{align}
where $p_i(t),v_i(t)\in\mathbb{R}$ are the longitudinal position and speed of vehicle $i$, and $u_i(t)\in\mathbb{R}$ is the control input.
We also impose the state and control constraints,
\begin{align}
    0 \leq v_{\min}&\leq v_i(t) \leq v_{\max}, \\
    u_{\min} &\leq u_i(t) \leq u_{\max},
\end{align}
where $v_{\min},v_{\max}$ are the minimum and maximum speed limit and $u_{\min},u_{\max}$ are the minimum and maximum control inputs.

In general, we have implemented optimal coordination, human-driven vehicles, and driver models in the IDS$^3$C \cite{chalaki2020experimental,chalaki2021CSM,Beaver2020DemonstrationCity}.
For simplicity, we present a scenario that employs the Intelligent Driver Model (IDM) \cite{Treiber2000}, which is known to mimic the behavior of human drivers.
This model outputs the acceleration for a vehicle $i$ based on the relative state of a preceding vehicle, $k\in\{1, 2, \dots, N\} \setminus \{i\}$,
\begin{equation} \label{eq:idm}
u_i(t) = u_{\max} \left[ 1 - \left( \frac{v_i(t)}{v_{\max}} \right)^\delta - \left( \frac{s^*(v_i(t),\Delta v_i(t))}{s_i(t)} \right)^2 \right],
\end{equation}
where $s^*$ is the desired headway of the vehicle,
\begin{equation} \label{eq:s_star}
s^*(v_i(t),\Delta v_i(t)) = s_0 + \max \left( 0, v_i(t) T + \frac{v_i(t) \Delta v_i(t)}{2 \sqrt{u_{\min} u_{\max}}} \right),
\end{equation}
where $s_i(t)$ is the bumper-to-bumper distance between vehicles $i$ and $k$, and $\Delta v_i(t) = v_i(t) - v_k(t)$. 
The constants $s_0, T, \delta$ are parameters that correspond to the standstill stopping distance, time headway, and an exponential factor that determine the acceleration and braking behavior, respectively.
Standard values for each of these parameters can be found in \cite{Treiber2000}.

We designed the roundabout scenario in Fig. \ref{fig:Roundabout} such that the two groups of vehicles would reach at the merging point at the same time.
To ensure safety, vehicles at the northern entry (Path $1$) must yield to roundabout traffic (Path $2$).
We achieved this by placing a virtual stopped vehicle at the position of the yield sign whenever a vehicle from Path $2$ was near the merging point \cite{Olfati-SaberFlockingTheory}.
When the area in front of the merging point was clear, the virtual vehicle was removed and vehicles on Path $1$ were allowed to enter the roundabout.
Otherwise, the vehicles traveling along Path $1$ form a queue and wait for the vehicles along Path $2$ to pass through the merging point.

\begin{figure}[ht]
    \centering
    \includegraphics[width=0.9\linewidth]{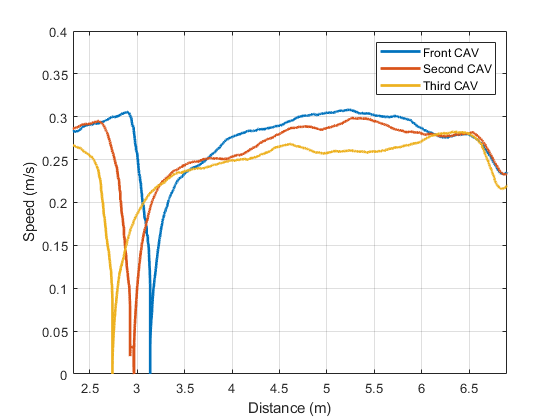}
    \caption{Speed vs position for the vehicles on Path $1$ in the IDS 3D City simulation with a $0.1$ s moving average filter applied.}
    \label{fig:simVel}
\end{figure}

\begin{figure}[ht]
    \centering
    \includegraphics[width=0.9\linewidth]{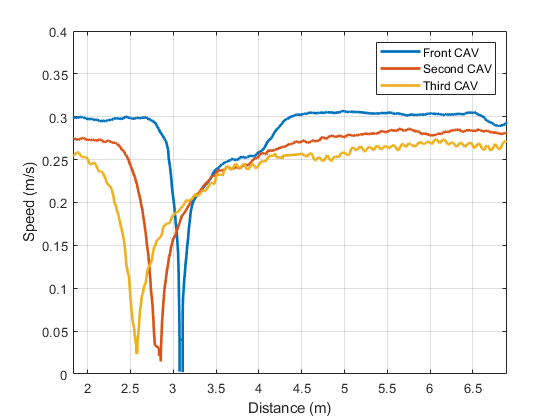}
    \caption{Speed vs position for the vehicles on Path $1$ in the IDS$^3$C experiment with a $0.1$ s moving average filter applied.}
    \label{fig:expVel}
\end{figure}

The speed of each vehicle following Path $1$ is plotted against position in Figs. \ref{fig:simVel} and \ref{fig:expVel} for the simulation and experiment, respectively.
The effect of the yield sign can be seen around $3.3$ m in both cases, where the front vehicle traveling on Path 1 comes to a full stop and a queue begins to form.
In simulation, after approximately two seconds, the front vehicle squeezes into a gap and merges with the vehicles on Path 2.
This causes the second vehicle on Path 1 to creep forward before coming to a complete stop again.
In contrast, the front vehicle in the experiment comes to a complete stop, is unable to merge, and a queue forms behind it.
This resulted in all vehicles on Path 1 yielding to all vehicles on Path 2 before entering the roundabout.
This observation demonstrates that while the IDM controller and vehicle dynamics behave similarly, the delays, noise, and disturbances in the physical experiment ultimately prevent the front vehicle from merging early in this particular scenario.

\begin{figure}[ht]
    \centering
    \includegraphics[width=0.9\linewidth]{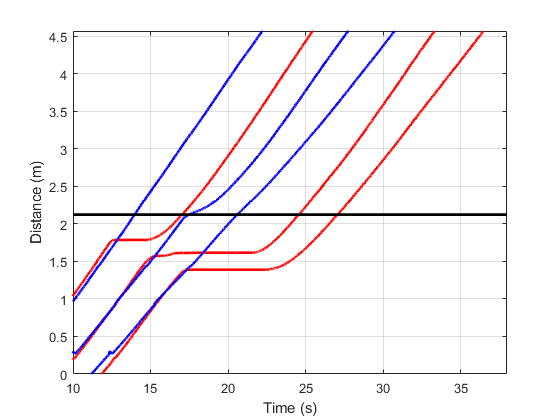}
    \caption{Position vs time trajectory vehicles in the simulation for the Path 1 (red) and Path 2 (blue). The horizontal black line corresponds to the position of merging point.}
    \label{fig:comparisonSim}
    \vspace{-8pt}
\end{figure}

\begin{figure}[ht]
    \centering
    \includegraphics[width=0.9\linewidth]{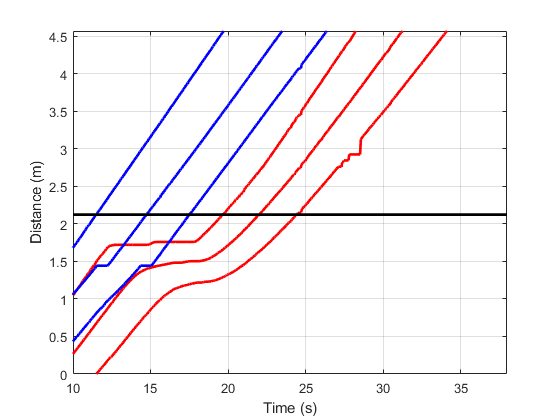}
    \caption{Position vs time trajectory vehicles in the experiment for the Path 1 (red) and Path 2 (blue). The horizontal black line corresponds to the position of merging point.}
    \label{fig:comparisonExp}
    \vspace{-8pt}
\end{figure}

The position of all vehicles is plotted against the time trajectory for the simulation and experiment in Figs. \ref{fig:comparisonSim} and \ref{fig:comparisonExp} respectively.
The horizontal black line around $2.1$ m marks one car length upstream from the merging point, and we have translated the reference frame of Path 2 such that the distance to the merging point is equal on both paths, i.e., the same distance corresponds to the same physical position.
Therefore, overlapping lines of different colors only correspond to collisions between vehicles at distances greater than $2.1$ m.
Despite the different vehicle dynamics in the simulation and experiment, Figs. \ref{fig:simVel} - \ref{fig:comparisonExp} demonstrate that both environments result in appropriate IDM behavior, and neither case leads to a collision between vehicles.
In addition, these results show that the simulated vehicles have smoother speed profiles compared to the experiment, as expected. 
Videos of the experiment and simulation, as well as supplemental material on the capabilities of the IDS $3$D City, can be found on our website \url{https://sites.google.com/view/ud-ids-lab/IDS3DCity}.

\section{Conclusion} \label{sec:conclusion}

In this work, we presented an overview of our virtual recreation of the IDS$^3$C.
Our simulation environment leverages the Unity game engine, AirSim, and ROS\# to control full-scale virtual vehicles, and to verify the behavior of our control algorithms before they are deployed in our physical environment.
We described how the simulated environment hooks into the control code for the physical city, which enables us to quickly iterate the design of an experiment and debug our control algorithms in simulation.
In particular, we illustrated that the intelligent driver model in a roundabout behaves properly, and we demonstrated that our control framework is independent of the underlying dynamics of individual vehicles.
Ongoing work includes performing experiments to implement our optimal control framework \cite{chalaki2020experimental} and mixed traffic input \cite{mahbub2021_platoonMixed}.
The most immediate direction for future research is to fully integrate the virtual vehicles with a physical experiment, resulting in an augmented-reality cyber-physical system.
Another intriguing direction is including AirSim drones in the simulated environment for applications that require air-ground cooperation.

\section*{Acknowledgements}
We would like to acknowledge Amanda Kelly for her help with building the virtual city environment.

\bibliographystyle{IEEEtran.bst} 
\bibliography{reference/IDS_Publications_04152022.bib, reference/ref.bib, reference/relatedPapers.bib}
\end{document}